\documentclass[conference]{IEEEtran}
\usepackage{cite}
\usepackage{amsmath,amssymb,amsfonts}
\usepackage{algorithmic}
\usepackage{graphicx}
\usepackage{dblfloatfix}
\usepackage{textcomp}
\usepackage{xcolor}
\usepackage{calc}
\usepackage{ifthen}
\usepackage{tikz,amsmath}
\usepackage{algorithmic}
\usepackage{algorithm}
\usepackage{comment}
\usepackage{eqparbox}
\usepackage{makecell}
\usepackage{adjustbox}
\usepackage[normalem]{ulem}
\newcommand{\tabitem}{~~\llap{\textbullet}~~}
\usepackage{pgfplots}
\usepackage{tikz}
\usepackage{xcolor}
\usepackage[normalem]{ulem}
\usepackage{pgfplots, pgfplotstable}
\usepackage{subcaption}

\definecolor{bblue}{HTML}{4F81BD}
\definecolor{rred}{HTML}{C0504D}
\definecolor{ggreen}{HTML}{9BBB59}
\definecolor{ppurple}{HTML}{9F4C7C}
\definecolor{yyellow}{HTML}{FFD700}
\definecolor{ppink}{HTML}{FE6F5E}
\definecolor{purpule}{HTML}{BF94E4}

\definecolor{white}{RGB}{255,255,255}
\definecolor{lightgray}{HTML}{C0C0C0}
\definecolor{gray}{HTML}{303030}
\definecolor{darkgray}{HTML}{333333}
\definecolor{blue}{HTML}{1DB1FC}
\definecolor{red}{HTML}{DC3912}
\definecolor{engineRed}{HTML}{E6194B}
\definecolor{cherryRed}{HTML}{e04136}
\definecolor{sherbertOrange}{HTML}{FF9900}
\definecolor{orange}{HTML}{f58231}
\definecolor{yellow}{HTML}{FFE119}
\definecolor{lime}{HTML}{BCF60C}
\definecolor{green}{HTML}{109618}
\definecolor{emerald}{HTML}{3CB44B}
\definecolor{cyan}{HTML}{46F0F0}
\definecolor{ocean}{HTML}{4363D8}
\definecolor{purple}{HTML}{9358FE}
\definecolor{grape}{HTML}{7748a3}
\definecolor{aquamarine}{HTML}{0C343D}
\definecolor{aqua}{RGB}{37,160,187}
\definecolor{jungleGreen}{HTML}{12b885}
\definecolor{grey}{HTML}{A4A4A4}
\definecolor{blueGrey}{HTML}{568cbf}
\definecolor{brightBlue}{HTML}{1566d1}
\definecolor{magenta}{HTML}{F032E6}
\definecolor{fuschia}{HTML}{a94cad}
\definecolor{orchid}{HTML}{911eb4}
\definecolor{bordeaux}{HTML}{800020}

\usetikzlibrary{patterns}
\definecolor{rosso}{RGB}{220,57,18}
\definecolor{giallo}{RGB}{255,153,0}
\definecolor{blu}{RGB}{102,140,217}
\definecolor{verde}{RGB}{16,150,24}
\definecolor{viola}{RGB}{153,0,153}

\usepackage{array}
\usepackage{fixltx2e}
\usepackage{dblfloatfix}
\begin{document}
\title{HyP-ABC: A Novel Automated Hyper-Parameter Tuning Algorithm Using Evolutionary Optimization}%{HyP-ABC:  Automated Hyper-Parameter Tuning Algorithm Using Modified Artificial Bee Colony Optimization}

\author{\IEEEauthorblockN{Leila Zahedi}
\IEEEauthorblockA{Knight Foundation School of \\Computing and Information Sciences\\
Florida International University\\
Miami, Florida 33199\\
Email: lzahe001@fiu.edu}
\and
\IEEEauthorblockN{Farid Ghareh Mohammadi}
\IEEEauthorblockA{Department of Computer Science\\
University of Georgia\\
Athens, Georgia, 30602\\
Email: farid.ghm@uga.edu}
\and
\IEEEauthorblockN{M. Hadi Amini}
\IEEEauthorblockA{Knight Foundation School of \\Computing and Information Sciences\\
Florida International University\\
Miami, Florida 33199\\
Email: amini@cs.fiu.edu}}


%Telephone: (800) 555--1212, Fax: (888) 555--1212}


\maketitle

\begin{abstract}
Machine learning techniques lend themselves as promising decision-making and analytic tools in a wide range of applications. Different ML algorithms have various hyper-parameters. In order to tailor an ML model towards a specific application, a large number of hyper-parameters should be tuned. Tuning the hyper-parameters directly affects the performance (accuracy and run-time). However, for large-scale search spaces, efficiently exploring the ample number of combinations of hyper-parameters is computationally challenging.  Existing automated hyper-parameter tuning techniques suffer from high time complexity. In this paper, we propose HyP-ABC, an automatic innovative hybrid hyper-parameter optimization algorithm using the modified artificial bee colony approach, to measure the classification accuracy of three ML algorithms, namely random forest, extreme gradient boosting, and support vector machine. Compared to the state-of-the-art techniques, HyP-ABC is more efficient and has a limited number of parameters to be tuned, making it worthwhile for real-world hyper-parameter optimization problems. We further compare our proposed HyP-ABC algorithm with state-of-the-art techniques. In order to ensure the robustness of the proposed method, the algorithm takes a wide range of feasible hyper-parameter values, and is tested using a real-world educational dataset.\\ %Results prove a significant improvement in tuning time and improved accuracy compared to previously suggested methods.\\
\end{abstract}


% \fi
\IEEEpeerreviewmaketitle

\section{Introduction}
\subsection{Motivation}
Deploying Machine Learning (ML) for real-world problems causes a number of challenges, e.g., selecting the proper model among a set of candidate models,   hyper-parameter tuning, and selecting the dataset's features to feed to ML models. An ML model's performance depends on such initial design decisions, which can be confusing to new users who want to choose the suitable model \cite{mohammadi2019parameter}.
These decisions need to be made based on some selection criteria, and are usually based on the model's obtained quality (performance indicator) \cite{neath2012bayesian}.
According to the principle of Occam's razor, a model should not be too simple nor too complex so that it can be efficient (in regards to computations) and at the same time capture data patterns, without overfitting \cite{biem2003model, mohammadi2019parameter}. Hyper-parameter tuning is choosing the parameters' values that have a more considerable impact on the ML model's final performance (e.g., accuracy and run-time). Hyper-parameters on an ML model control the convergence of the learning process. Hence, to find an ML models' optimal performance, it is crucial to find the optimal values for hyper-parameters  \cite{claesen2015hyperparameter}.
Independent from the ML users, obtaining desirable results for their specific dataset  can be conducted manual and tedious \cite{sharmaart, yang2020hyperparameter}. This is where Automated ML (AutoML) comes into the picture, to alleviate such burden in an automated way. Automated Hyper-parameter Optimization (HPO) is one of the decisive and primary tasks of AutoML \cite{sui2020bayesian}.

In HPO, the challenge is that there exists a large number of possible configurations. However, there is always a trade-off between performance and time. Therefore, there should be a balance in the number of candidates and the resources allocated to them \cite{yang2020hyperparameter}. Population-Based Algorithms (PBAs) have recently gained much attention across fields as they are used for different applications. Swarm intelligence (SI), a type of PBA, is the collective behavior of self-organized and decentralized swarms (such as birds, ants, or bees). The self-organization and division of labor characteristic of swarm intelligence can be observed in bee colonies, so they are ideal for developing intelligent approaches. Among different bee colony algorithms, Artificial Bee Colony (ABC) is the most popular due to its capabilities \cite{karaboga2014comprehensive}. Therefore, in this work, ABC is chosen to tackle large-scale optimization problems (regarding search space) in HPO.

 

%========================================
\subsection{Objective}
\label{sec:ResearchObjective}
%========================================
Manual search and automated exhaustive search among \begin{math}n^{k}\end{math} configurations of hyper-parameters are impractical and time inefficient \cite{yin2016classification}. Manual tuning does not provide reproducibility. Exhaustive search, on the other hand, suffers from dimensionality issues in large search spaces. As such, there has been increased research in HPO to optimize the performance of ML models regarding both accuracy and time.

When making decisions, time is of the essence. Hence, many of the black-box optimization models are not a good fit for such optimization problems because they do not consider the function evaluation time \cite{yang2020hyperparameter}. Therefore, proper algorithms should be applied to such a problem to find the optimal set of hyper-parameters. Mohammadi et al. provided different examples of applying PBAs, such as feature extraction, in different domains and encouraged researchers to apply such methods to solve large-scale optimization problems \cite{mohammadi2020evolutionary, mohammadi2020applications}. 

On the other hand, automated HPO methods may change the final model in comparison to when a model is being trained with default hyper-parameters \cite{zahedi2021search}. Hence, they also provide fairness to research and scientific studies. Our goal is to design and develop an HPO framework to tackle the challenges discussed above in this work. The framework utilizes ABC to minimize the run-time caused by large dimensions of search-space scales, known as Curse of Dimensionality (CoD), and possibly improving the accuracy of the ML model. This framework will assist the predictions for intelligent and real-time educational decision-making and lessen the costs and manual and human efforts.

\begin{figure}
	\centering
		\includegraphics[width =0.45 \textwidth]{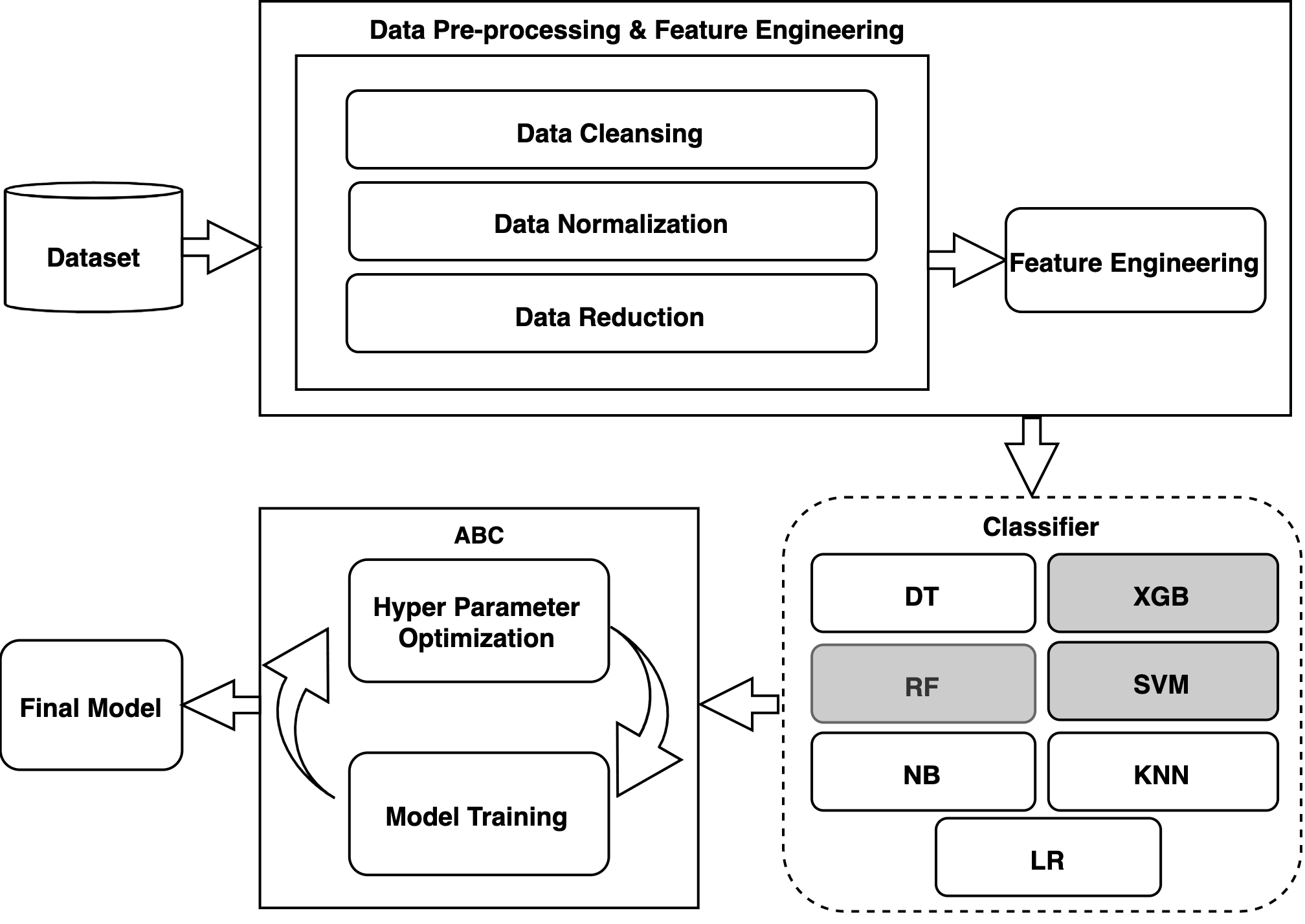}
	\centering
	\caption{General framework of the study}
	\label{fig:general}
\end{figure}

In the literature, ABC has been used to tune the parameters of deep learning algorithms, convolutional neural networks, and least squares support vector machine \cite{badem2018new, ozcan2020human, ozcan2019transfer}. In this paper, a real-world educational dataset is used to build a binary classification model to predict students' success, specifically graduation. Figure \ref{fig:general} shows the general diagram of our experiment. We utilize HyP-ABC to tune the main hyper-parameters of Random Forest (RF), Extreme Gradient Boosting (XGBoost), and Support Vector Machine (SVM). These three models are based on the reports from an earlier study \cite{zahedi2021search} indicating the high tuning-time using exhaustive automated HPO approaches, namely Grid Search (GS) and Random Search (RS). 

\subsection{Contribution}
\label{sec:contribution}
%========================================
The main contributions of this study are as follows:
\begin{enumerate}
    \item This paper is developing a novel optimization method for three ML algorithms using a modified heuristic optimization algorithm.
    \item The propsoed HyP-ABC algorithm outperforms the state-of-the-art HPO methods utilized in an existing study in 2021 \cite{zahedi2021search}, in terms of both tuning time and accuracy.
    \item This paper is the first to investigate the optimization of  ML models tailored towards the MIDFIELD dataset.% using a heuristic optimization algorithm.
    \item Unlike most of the previous works that try to address the issues on single ML models, in this work, we explore utilizing the HyP-ABC on three different ML algorithms to tune their hyper-parameters and select the final model among them based on their performance.
\end{enumerate}
\subsection{Organization of Paper}
\label{sec:Organization}
The rest of the paper is organized as follows: in section \ref{sec:PBO} we provide an introduction to PBA and a comprehensive explanation of ABC, in section \ref{sec:PBO}. Then, we present the related work regarding hyper-parameter tuning and different automated tuning methods in section \ref{sec:related}. We also cover ABC applications in this section. Section \ref{sec:HPO-ABC} covers hyper-parameter tuning using the ABC approach and the advantages of ABC over other techniques. In section \ref{sec:MABC}, we propose the HyP-ABC algorithm, which is an ABC-based algorithm fitted to HPO problems. The experimental methodology, search space, and dataset are then presented in section \ref{sec:methodology}. Section \ref{sec:results} presents experimental results to demonstrate the performance achievable by the proposed approach. Last, section \ref{sec:conclusion} concludes the paper and discusses future directions.

\section{Population-Based Optimization Algorithms}
\label{sec:PBO}
Population-based optimization algorithms work based on generating and updating individuals in each generation; this continues until the final optimal solution is identified \cite{yazdani2021survey}. These algorithms are based on the interaction between different individuals called food sources to find the near-global optimal solutions. These algorithms garnered much attention because of their excellent performance and are particularly useful to solve optimization problems \cite{pierezan2018coyote}. PBAs include different heuristic algorithms such as EAs, Genetic Algorithms (GAs), Particle Swarm Optimization (PSO), Ant Colony Optimization (ACO), Grey Wolf Optimization (GWO), Fish Swarm Algorithms (FSA), and Artificial Bee Colony (ABC). The main difference between these algorithms is how a population is generated and selected \cite{yao2018taking}. One of the main advantages of these algorithms is that unlike many model-based optimization methods (such as Bayesian optimization), they have the capability of parallelization \cite{hutter2019automated}. Among these algorithms, ABC is one of the most popular methods due to the excellent exploration and exploitation capabilities to find the optimal solution \cite{karaboga2007artificial}. Therefore, ABC was chosen for further exploration in this study to answer HPO challenges.

\subsection{Artificial Bee Colony (ABC)}
ABC, first defined by Karaboga \cite{karaboga2007powerful}, is one of the most recent and popular swarm intelligence algorithms that simulate honey bees' foraging behavior. In ABC, different groups of bees fly around in an area (search space). The bee colony, in this algorithm, is divided into three groups: 1) employed, 2) onlookers, and 3) scouts.

At the initialization stage, a random set of food sources get selected by \textbf{scout bee}, and their amount of nectar is determined. Then scout bees share the nectar information with employed bees. Then the \textbf{employed bees} visit those food sources and choose a new food source in that neighborhood and compare with the current food source. Next, \textbf{onlooker bee} may prefer a food source based on the information employed bees shared with them (nectar amount). The higher the amount of nectar around food increases the chance of getting selected by onlooker bees. Then the onlooker bee chooses a new food source in that neighborhood and makes the comparison. The better food source gets replaced in both employed and onlooker phases. When the bees abandon a food source, the scout bee selects a random food source and gets replaced with the left food source.\\
Employed and onlooker bees both do the \emph{exploration} based on their own experience, leave poor food sources, and move toward better ones through a greedy selection process. However, the difference between these two phases is that a food source may or may not be selected in the onlooker phase. If the food source is not selected, the onlooker bee moves to the next food source, while in the employed phase, each food source is assigned to one employed bee. When a food source exploitation is exhausted and can not be further improved (based on trial limits), the employed bee discards the food source and becomes a scout bee and start \emph{exploration}.

The pseudo-code of the original ABC is described in Algorithm \ref{alg:ABC} \cite{karaboga2009comparative}, to show the main steps of the ABC approach.

  \begin{algorithm}[H]
  \caption{Implementation of original ABC}
  \begin{algorithmic}[1]
  \renewcommand{\algorithmicrequire}{\textbf{Input:}}
  \renewcommand{\algorithmicensure}{\textbf{Output:}}
  %\ENSURE  Best food source identified 
  \STATE {Initialize population}
  \WHILE{Stopping criteria is not met}
  \STATE {Assign food sources to employed bees}
  \STATE Place the onlooker bees on the food sources based on the amount of nectar
  \STATE Send the scouts to different areas to find new food sources
  \STATE Memorize and update the best food source
  \ENDWHILE
  \RETURN Best found food source
  \end{algorithmic}
  \label{alg:ABC}
  \end{algorithm}
  
\section{Related Works}
\label{sec:related}

\subsection{Hyperparameter Tuning}
\label{sec:hyperparameter}
GS is a decision-theoretic and brute-force method that searches all the hyper-parameters within a fixed search space. The advantage of this method is that it obtains the optimal solution in a discrete search space \cite{zahedi2021search}. However, it is computationally expensive in large-scale spaces.

RS is also a decision-theoretic approach that randomly selects the configurations with limited resources (time or number of iterations). This method works well for search spaces containing continuous search spaces. However, luck plays a part in this method, and given more resources increases the chance of getting better results \cite{zahedi2021search}.

Unlike GS and RS, Bayesian Optimization (BO) \cite{eggensperger2013towards} prevents evaluating many of the unnecessary configurations based on the evaluations of the previous steps. However, since BS methods work sequentially to balance the exploration and exploitation, they are complicated to parallelize.

For some models, such as tree-based models (RF and XGB), the number of main hyper-parameters and their ranges are higher than other ML models. This leads to large search space scales, making them the most complex ML models to be tuned \cite{hutter2019automated}. Furthermore, since different ML algorithms have different types of hyper-parameters (continuous, integer, and categorical), they should be treated differently in tuning processes \cite{decastro2019effect}. Hence, they are the main focus of this study. 

For the ML models in this study, ABC is selected. It has a reasonable convergence rate and enables parallel executions to improve tuning time, particularly for models that often require massive training time. Some other techniques, like GA, can also be used, but they may cost more time than ABC since it is difficult to parallelize these techniques. Also, techniques such as PSO have a reasonable convergence rate; however, the PSO technique has a higher chance of sticking into local optimums.

\begin{table*}[ht]
\centering
\caption{Comparison of GA, PSO and ABC algorithms}
\label{tab:HPOcompare}
%\begin{adjustbox}{width=\columnwidth,center}
\def\arraystretch{1.5}
\begin{tabular}{|c|c|c|c|}
\hline
\textbf{Method} & \textbf{Advantages}
& \textbf{Disadvantages}   & \textbf{\makecell{Time Complexity}}\\
\hline
\textbf{GA}   &   \makecell{\tabitem No proper initialization is needed}   & \makecell{\tabitem Poor with parallelization\\ \tabitem Introduces new parameters\\ \tabitem Slow Convergence}   &  \makecell{\begin{math}O(n^{2})\end{math}}\\
\hline
\textbf{PSO} &   \makecell{\tabitem Enables parallelization \tabitem Faster convergence}  &  \makecell{\tabitem Needs proper initialization\\\tabitem May stuck in local optimum}   & \makecell{\begin{math}O(n\log{n})\end{math}}  \\
\hline
\textbf{ABC} &   \makecell{\tabitem Enables Parallelization\tabitem Higher efficiency \\\tabitem Balances exploration \& exploitation \\\tabitem More likely to find global optimum}  &  \makecell{\tabitem Need proper initialization \\\tabitem Slower convergence than PSO}   &  \makecell{\begin{math}O(n)\end{math}}  \\ % O(3n*d*I_{max})
\hline
\end{tabular}
%\end{adjustbox}
\end{table*}

\subsection{ABC Applications}
\label{sec:relatedABC}
In this section, we cover some of the previous studies of ABC for different optimization problems. one of the important trend that ABC started in \cite{mohammadi2014image} by proposing IFAB, an approach to apply ABC to distinguish between clean and unclean images. The authors technically applied ABC by converting the ABC into a way it works best for discrete problems. Moreover, Sarac Essiz and Oturakci developed a comparative analysis. They explored the effects of the ABC-based feature selection algorithm \cite{xu2020duplicationsha} to eliminates non-informative features on the classification performance of a cyberbully detection problem. They also showed that their method has a better performance than some conventional methods such as information gain, Relief, and chi-square \cite{sarac2021artificial}.% In another study, Zhu et al. proposed a neuro-evolution framework based on ABC on MNIST dataset to automatically design and evolve hyper-parameters of Convolutional Neural Networks (CNNs) \cite{zhu2019evolutionary}. 

Amar and Zeraibi \cite{amar2018application}, combined ABC and Support Vector Regression (SVR) to predict minimum miscibility pressure and improve the oil recovery process. In their study, ABC was used to find the best hyper-parameters of the SVR model. In another study, Dokeroglu et al. developed a hybrid ABC optimization algorithm for the quadratic assignment problem using Tabu search to simulate exploration and exploitation phases  \cite{dokeroglu2019artificial}. The results showed that an ABC performs well for most quadratic assignment problems and can compete with other state-of-the-art meta-heuristic algorithms existing in the literature.

Choong et al. utilized a modified choice function for the ABC algorithm. This algorithm adjusts the neighborhood search performed in the employed and onlooker phases. The results showed that the modified choice function brought advantages to the search process \cite{chang2018solving}. Mohammadi et al. provided a comprehensive review of evolutionary-based image processing on real-time processing in universal image steganalysis and detected hidden messages in images. Since training ML models is time-consuming, the authors argued that the use of algorithms such as the ABC algorithm is of great benefit to solve NP-hard problems caused by CoD \cite{mohammadi2019evolutionary}.
In \cite{mala2021hybrid}, the authors proposed two algorithms to modify two steps of the original ABC. The first algorithm was to employ neural network initialization, and the second algorithm utilized stochastic gradient descent in the employed phase of ABC to improve the convergence. The authors examined the performance of the modified ABC on three benchmark datasets and observed promising results.

In another study, Zhao et al. performed a comparative analysis that proposed a modified version of ABC to improve the performance of SVM classification using parameter optimization. The proposed method utilized chaotic sequences for the initialization step. They also defined an adaptive step size for the neighborhood search to boost the algorithm convergence. The modified ABC algorithm was examined to perform classification on two hyper-spectral images. The authors then compared the results with three other PBAs and observed the superiority of the ABC method over the rest of the algorithms regarding both performance and efficiency \cite{zhao2020improvement}.

Pandiri and Singh \cite{pandiri2018hyper} developed a hyper-heuristic-based artificial bee colony algorithm for the k-Interconnected multi-depot multi-traveling salesman problem (k-IMDMTSP). The authors stated that due to the parameter values, it is impossible to have a unique algorithm that outperforms the rest of the algorithms emanating from k-IMDMTSP. Hence, they leveraged a hyper-heuristic-based ABC algorithm for this problem. The authors presented an encoding scheme to be used inside the algorithm. The experimental results for this study exhibited smaller search space than previous encoding schemes, yet they performed better than other approaches existing in the literature regarding quality and run-time.

In another study, Mazini et al. took advantage of the parameters regulation method of ABC for feature selection to propose a reliable hybrid method to detect anomaly network-based intrusion. Experimental results on an unbalanced network traffic dataset exhibited significant improvement in performance and detection rate compared to other intrusion detection systems in various scenarios \cite{mazini2019anomaly}. Gunel and Gor developed a modified ABC algorithm to solve initial value problems. To define a mutation operator, the authors in this study used a dynamically constructed hyper-sphere to generate new food solutions; This method improved the exploitation ability of ABC. In this approach, the best solution was used to define the mutation operator. The solutions for the differential equations were yielded through training neural networks applying the modified ABC \cite{gunel2019modification}.

In \cite{sayed2019parameters}, Sayed et al. used an ABC-based approach to tune the hyper-parameter of ABC to improve the learning performance. In this study, the authors adjust the main hyper-parameter of the SVM model to enhance the accuracy. The proposed method was examined on six different datasets and was also compared with popular swarm algorithms. The experimental results showed promising results compared with the mentioned algorithms.
Agrawal also proposed an extended optimization of ABC using some features of the Gaussian ABC scheme to tackle some of the disadvantages of the original ABC, such as a slow convergence. The proposed method was used to adjust the exploration and exploitation phases. They examined the proposed approach on some benchmark datasets and observed that the proposed method outperformed the original ABC in most experiments \cite{agrawal2020modified}.

\section{Hyper-parameter Optimization using ABC}
\label{sec:HPO-ABC}
There are different optimization problems in multiple fields. To such problems, there exist classical approaches and heuristic approaches. Classical techniques are not efficient enough in solving optimization problems. This is primarily due to dimensionality. Heuristic approaches such as genetic algorithms and evolutionary algorithms do not suffer from many of the drawbacks of classical methods. ABC is a stochastic, population-based optimization technique and belongs to the family of swarm intelligence techniques. ABC is inspired by social interactions in honeybees and has recently garnered much success in different applications. It is as simple as some other swarm techniques such as PSO and is beneficial for NP-hard problems. Looking for the best hyper-parameter in the hyper-parameters search space is also an NP-hard problem with a time-complexity of \begin{math}O(n^{k})\end{math}. Among heuristics, ABC has shown more encouraging behavior. Hence, ABC is being explored in this study to find the best hyper-parameters of ML algorithms. This study was undertaken specifically to tackle the challenge regarding the high tuning time authors mentioned in an earlier work \cite{zahedi2021search} regarding models such as RF, XGBoost, and SVM.

\label{sec:ABC}
\begin{figure*}[b]
	\centering
		\includegraphics[width =0.60 \textwidth]{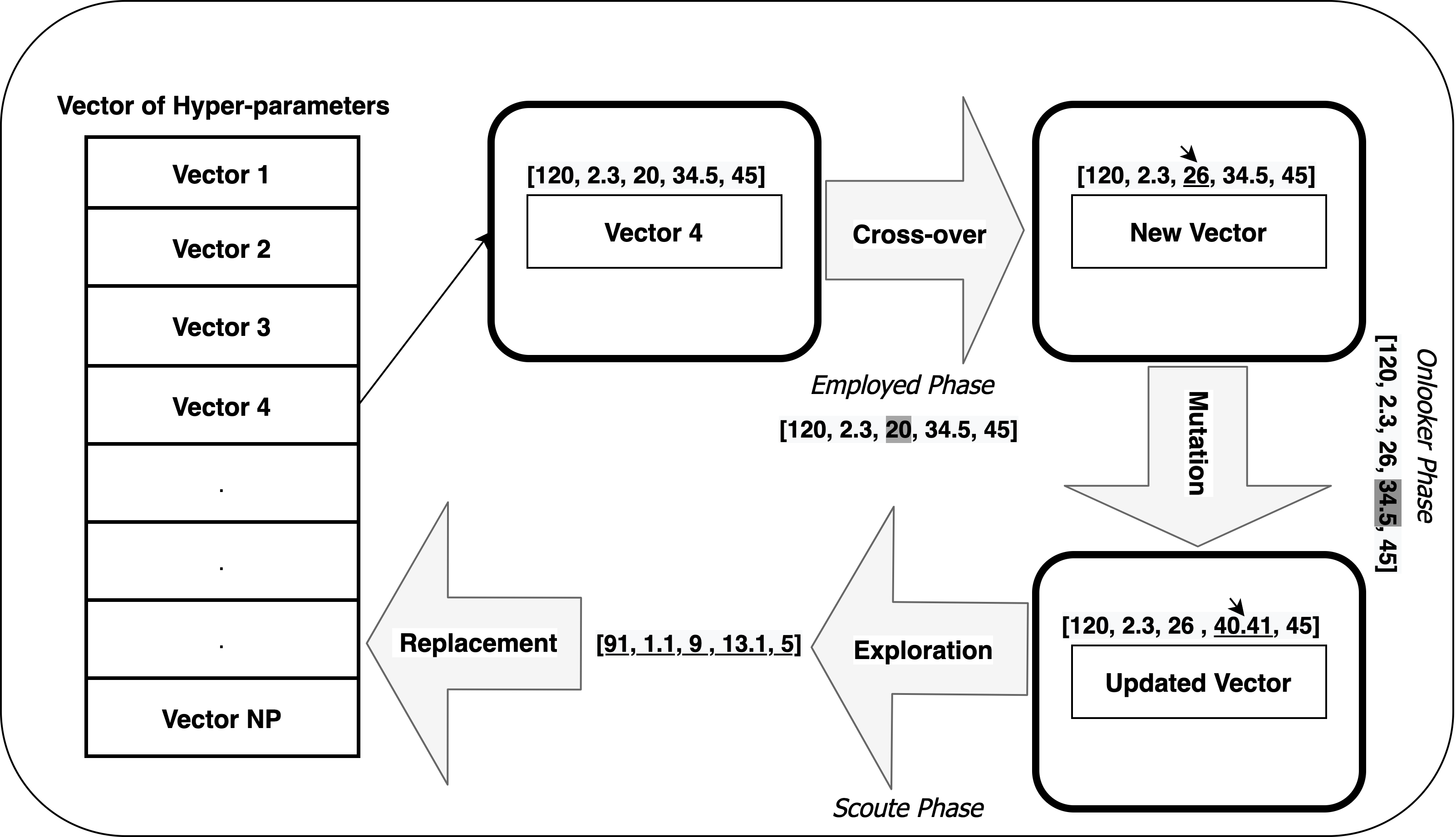}
	\centering
	\caption{General scheme of the HyP-ABC}
	\label{fig:POA}
\end{figure*}

\subsection{Time Complexity}
\label{sec:time}
Table \ref{tab:HPOcompare} summarizes each of these methods' advantages and disadvantages along with their time complexity. We cover the time complexity of our method in the next section.

As shown in Table \ref{tab:HPOcompare}, the time complexities of the methods mentioned in the previous section are summarized. ABC requires a one-pass scan for each of the initialization, employed, and onlooker phases. Also, the ABC algorithm can repeat based on a maximum number of iterations defined for the algorithm. The time complexity for each of the neighborhood searches is \begin{math}O(N*D)\end{math} where N is the number of population, and $D$ is the dimensionality. Therefore, the complexity of ABC can be presented as \begin{math}O(3N*D*I_{max})\end{math} where $I_{max}$ is the expected maximum number of iterations. Since $D$ and $I_{max}$ are constant the overall time complexity of the HyP-ABC is linear (\begin{math}O(N)\end{math}). Hence, this research is based on an ABC-based approach.

\subsection{Advantages of ABC}
BO \cite{eggensperger2013towards}, GA \cite{lessmann2005optimizing} and PSO \cite{guo2008novel} are among the common approaches used for HPO problems.
There are various features that ABC has which make it efficient in solving optimization problems, especially for hyper-parameter tuning are as follows:
\begin{enumerate}
\item The underlying concept and implementation are easy and offer high accuracy.
\item Although almost all meta-heuristic make use of randomization to have global research as well as local search, ABC balances the exploration and exploitation steps (local and global search) \cite{karaboga2009comparative}. This enables the algorithm to search for various parts of the search space. Randomization also helps the model not get stuck in the local minimum and complete the global search. This is despite BO or PSO methods that may stick to a local optimum and which fail to reach a global optimum \cite{yang2020hyperparameter}.
\item Parallelization is one the advantages of PBA because each population can be assessed on one machine \cite{hutter2019automated}, while sequential methods such as BOs and GAs are challenging to parallelize since solutions are dependant on each other \cite{decastro2019effect}.
\item Some PBA methods such as GA have some additional hyper-parameters (crossover and mutation probability, population size, number of generations, crossover, mutation, and selection operators) to tune; therefore, GA has lower convergence speed \cite{lessmann2005optimizing}. Having fewer parameters to be tuned by the user is one of the advantages of methods such as PSO and ABC.
\end{enumerate}

\subsection{General objective of study}
Many optimization problems, including hyper-parameter tuning of ML algorithms, all or some of the hyper-parameter are discrete. The schematic objective of HyP-ABC is shown in Figure \ref{fig:POA}. 
As shown in Figure \ref{fig:POA}, the vectors of hyper-parameters could consist of discrete (integer and categorical) and/or continuous (float) values. Solving these problems is even more challenging than the problems with only continuous variables. Since the basic ABC is only applicable to continuous problems, proper strategies should be adopted to make them applicable to discrete or combinatory problems. Therefore we need to modify the original ABC to adopt it to HPO problems.

The first step is generating the initial population. Each sample should be described as a vector, including a set of hyper-parameters. This step is done by a {\textbf{scout bee}}. Corresponding to each of the vectors, we have an objective function produced after training the ML model (accuracy of the model) and a fitness function. The fitness function is calculated based on the objective function. After the population is generated, each vector is assigned to one \textbf{employed bee}. The employed bee generates a new vector of hyper-parameters in the neighborhood, meaning that only one random hyper-parameter of the current vector changes based on some functions being applied on the same hyper-parameter of another vector (neighbor) in the population. If the new vector has a better fitness, it gets replaced with the current vector. 

In the mutation step, \textbf{onlooker bees} start operating. They calculate the quality of each vector. Each of these vectors may or may not get selected. The higher the quality of the vector, the higher the probability of it being selected. If a vector is selected, the onlooker bee generates a new vector of hyper-parameter in the vicinity and starts assessing the new vector by comparing its fitness with the previous vector. If the new vector has a better fitness, it gets replaced with the previous vector (similar to the employed bee phase).

In the next step, an exhausted vector of hyper-parameters (the vector that had the chance to improve the fitness but did not) is selected and replaced with a random vector generated by the scout bee. Unlike the vectors generated in employed and onlooker phases, the new vector consists of randomly generated values for all hyper-parameters. This phase helps the algorithm not to stick to the local optimum. If stopping criteria are met at any of the above phases, the algorithm stops and returns the optimal performance of the ML model, and if not, the process repeats until stopping criteria are met.

\section{HyP-ABC: Modified ABC Algorithm}\label{sec:MABC}
\subsection{Converting continuous ABC to combinatory version}
As we mentioned in the previous section, we face a combination of hyper-parameters' types while original ABC is only applicable to continuous problems. Also, the range of variables in basic ABC is the same for all the dimensions. Therefore proper strategies should be adopted to make it applicable to discrete or combinatory problems. There are different strategies used for tackling discrete optimization in swarm optimization \cite{ziari2012integrated, wu2008feeder,durgut2017artificial,chang2018solving}. Rounding off is one of the most common approaches to tackle discrete variables. In this approach, the integer variables are treated similarly as continuous variables during the optimization process.

However, in most studies, when the optimization process is done, the variables are rounded off to the nearest integer number. Simplicity and low computational cost are among the main advantages of this method. However, entering impossible regions and high variation of fitness value of rounded value and the original value are among the disadvantages of this method. In this study, we use a function to adjust each vector's member based on their type and in each iteration. Hence, we yield precise accuracy with the very same hyper-parameters at the end of the optimization process.

Hyper-parameters have different types (discrete, categorical, continuous), and therefore should be treated differently in each iteration of generating food sources. Instead of using strings for categorical variables in the population generation phase, we encode them and generate integer numbers (and treat them as integer variables afterward). We then convert them again to corresponding strings after the optimization process is done and just before training the model.

From this step on, if the selected hyper-parameter's type by employed/onlooker is an integer (or categorical), the newly generated food source is modified to an accepted vector by converting that hyper-parameter to the closest integer number. For continuous variables, the algorithm performs similarly to the original ABC. Algorithm \ref{alg:hypabc} shows the steps for the HyP-ABC we applied in this study.

We also added an additional step to check if the updated food source equals the current food source. This is specifically useful when the categorical hyper-parameter is the variable selected to be changed due to having fewer options. Hence the HyP-ABC prevents training the ML model with the same hyper-parameters and tries to find a different food source, and the objective function is not calculated for a configuration that is already evaluated. For binary categorical variables (such as criterion in RF), we use the XOR operator to flip the value of the hyper-parameter.
 \begin{algorithm}[H]
  \caption{Hybrid artificial ABC algorithm}
  \begin{algorithmic}[1]
  \renewcommand{\algorithmicrequire}{\textbf{Input:}}
  \renewcommand{\algorithmicensure}{\textbf{Output:}}
  \REQUIRE $Population\_number$, $Search\_Space$
  \ENSURE  Best food source identified\\
  \STATE Call a function to create an initial population X{i} based on the hyper-parameters type and range
  \STATE Set the trail=0 for all food sources
  \WHILE{Stopping criteria is not met}
  \FOR{ $i$ to $Population\_number$}
  \STATE Employed bee generate a new food \(N{i}\) source in neighborhood and modify the new food source to an accepted food source \(M{i}\) (accepted type within range)
  \IF{ \(M{i}\) $==$ \(X{i}\) }
  \STATE Move to step 5
  \ENDIF
  \STATE Train the ML model with the modified food source
  \IF{ \(f(M{i})\) $<$ \(f(X{i})\) }
  \STATE \(X{i}\) = \(M{i}\)
  \STATE Reset trial
  \ELSE
  \STATE Increment trial
  \ENDIF
  \ENDFOR
  \FOR{ $i$ to $Population\_number$}
  \IF{ \((rand{(0,1)})\) $>$ \((P{i})\) }
  \STATE Onlooker bee generate a new food source in neighborhood \(N{i}\) and then modify the new food source to an accepted food source \(M{i}\)
  \IF{ \(M{i}\) $==$ \(X{i}\) }
  \STATE Move to step 5
  \ENDIF
  \STATE Train the ML model with the updated food source
  \IF{ \(f(N{i})\) $<$ \(f(X{i})\) }
  \STATE \(X{i}\) = \(M{i}\)
  \STATE Reset trial
  \ELSE
  \STATE Increment trial
  \ENDIF
  \ELSE 
  \STATE Onlooker disregard the food source and move to the next food source
  \ENDIF
  \ENDFOR
  \STATE Memorize and update the best food source
  \IF {trial $>$limit}
  \STATE Scout bee generates a new food source
  \ENDIF
  \ENDWHILE
  \RETURN Best found food source
  \end{algorithmic}
  \label{alg:hypabc}
  \end{algorithm}

In all phases, values outside the ranges defined for the hyper-parameters get replaced with lower bound or higher bound values before training the model. Stopping conditions in this algorithm are when the desired accuracy is reached or when the algorithm evaluated a specific number of evaluations. In this experiment, we set the maximum number of evaluations equal to the number of iterations in RS to have a fair comparison of HPO methods.

\subsection{HyP-ABC Steps}
For the random initialization of the population. Each food source is a vector of hyper-parameters, $X_i$, with the length of $D$, where $D$ is the dimension of vector or the number of hyper-parameters. Unlike original ABC, in HyP-ABC, each vector member may have discrete or continuous type:
\begin{equation} X_{i,j}=x_{min,j}+rand(0,1)(x_{max,j}-x_{min,j}) \end{equation}
Where $X_{i,j}$ is each element in the vector and $x_{max,j}$ and $x_{min,j}$ are upper-bound and lower-bound for a specific hyper-parameter ($j$). Then, each vector is assigned to an employed bee. The employed bee generate a new food source $N_{i,j}$ in that neighborhood by changing only one of the hyper-parameter using below formula:
\begin{equation} N_{i,j}=x_{i,j}+rand(-1,1)(x_{i,j}-x_{k,j}) \end{equation}
Where $x_{k,j}$ represent a value in the neighborhood of current food source and for the same dimension. Therefore $k$ and $i$ are different. Then the fitness of the $N_{i}$ and $X_{i}$ are compared. In case the fitness of $N_{i}$ is better, it get replaced with $X_{i}$ and therefore becomes a new member of the population. Fitness is calculated from the objective function (classification accuracy):
\begin{equation}
fit_{i}=
\begin{cases}
\frac{1} {1+ f_{i}}, & \text{$f_{i}$ $\geq$ 0} \\
{1+ abs(f_{i})}, & \text{$f_{i}$ $<$ 0}
\end{cases}
\end{equation}
When employed bees complete the search, onlooker bees receive information from employed bees and may select a vector based on selection probability values. The vector with higher quality (higher selection probability) has a better chance to get selected by onlooker bees. In this phase, roulette wheel selection, as shown in the below formula, is utilized to calculate the probabilities.
\begin{equation} P_{i}=\frac{f_{i}}{\sum_{j=1}^{PN}(f_{j})} \end{equation}
Where $f_{i}$ is the fitness value for the \(i\)th food source and $PN$ is the population number. Next, scout bees start explorations searching random configurations of search space without using experience or memorizing the locations. They do not use greedy selection when exploring for new solutions. Exploration by scout bees is through below formula:
\begin{equation} X_{i,j}=x_{min}+rand(0,1)(x_{max}-x_{min}) \end{equation}

\section{Experimental Methodology}
\label{sec:methodology}

In this section, we describe our methodology for this study's experiment. The process is performed in several consecutive phases. It includes data pre-processing, feature engineering, leveraging ML models, and optimization steps. The Hyp-ABC algorithm proposed in this study is replaced with the HPO methods utilized in our most recent work in 2021 \cite{zahedi2021search}. 

\subsection{Data pre-processing and feature Engineering}
Data pre-processing is a technique used to transform the raw data into an understandable format for ML algorithms. It includes data cleansing, data normalization, and data reduction. For data cleansing, we removed the features with more than 60\% percent of the values missing. Data normalization is also a method to standardize the data when the different features' values vary widely. \textit{StandardScaler} from Scikit-learn library was used for scaling the input data. We filtered the data in the data reduction step and included only the target population, students from computing fields.

In the feature engineering step, one-hot encoding -encoding categorical variables to binary variables for each unique category- is used.

\subsection{Hyper-parameter tuning}
\begin{figure*}
	\centering
	\includegraphics[width =0.750\textwidth]{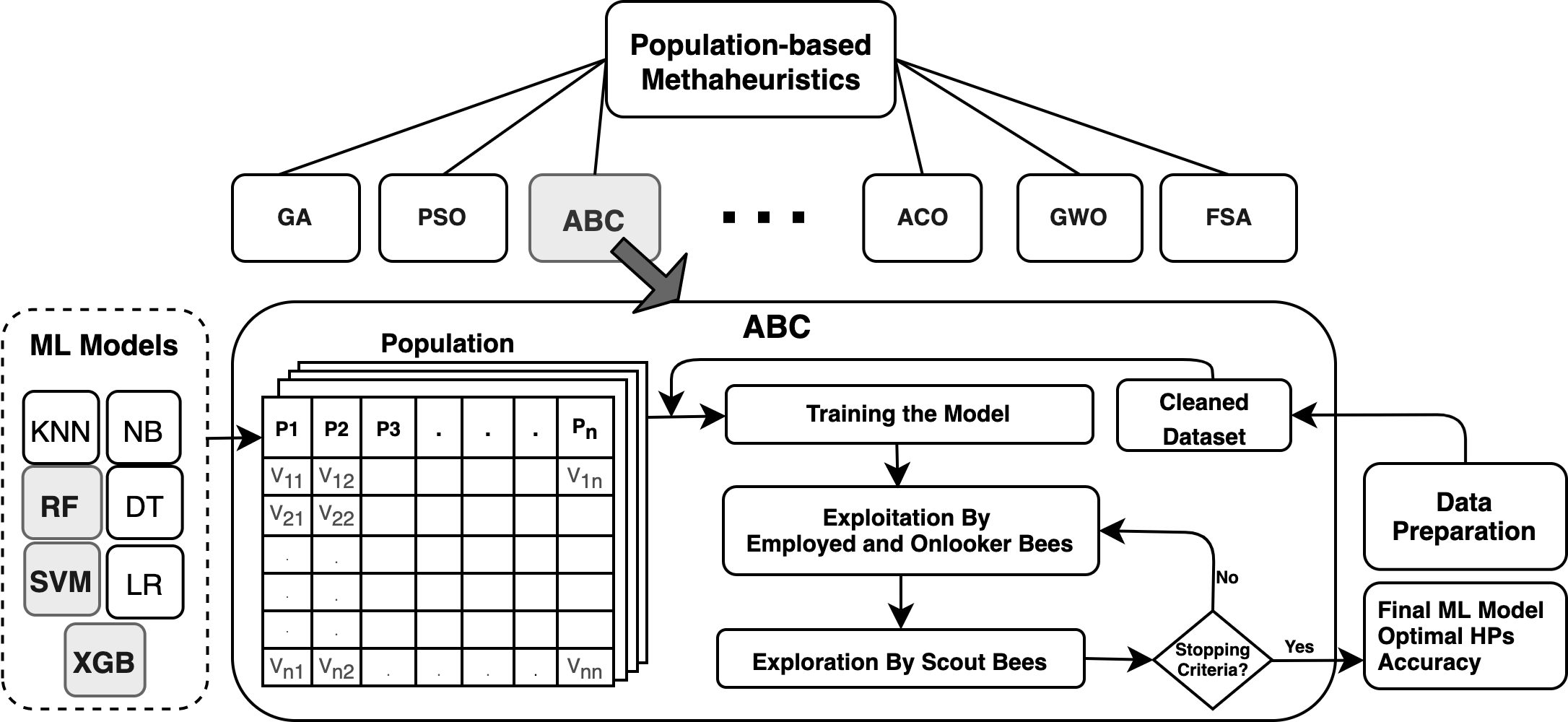}
	\centering
	\caption{The schematic process of the study}
	\label{fig:ABC}
\end{figure*}

Tuning hyper-parameters is essential to yield the best performance of an ML model. Due to the varied nature of the hyper-parameters in different ML models, we implement a HyP-ABC to tune the hyper-parameters of three ML models.
HyP-ABC is replaced and compared with an automated HPO method performed in our most recent study in 2021 \cite{zahedi2021search}. Algorithm \ref{alg:GRS} shows the previous work process, along with the changes we made to improve the tuning process regarding run-time and performance. Figure \ref{fig:ABC} shows the overall schematic process of our framework.

\newcommand\redsout{\bgroup\markoverwith{\textcolor{red}{\rule[0.5ex]{2pt}{0.4pt}}}\ULon}
  \begin{algorithm}[b]
  \caption{Automated optimization in a previous work \cite{zahedi2021search}} 
  \begin{algorithmic}[1]
  \renewcommand{\algorithmicrequire}{\textbf{Input:}}
  \renewcommand{\algorithmicensure}{\textbf{Output:}}
  \REQUIRE \(List_{ML}\) of models (DT, RF, NB, LR, XGB, SVM, KNN) \& raw dataset
  \ENSURE  Best model along performance and optimal architecture 
   \STATE Call PreProcessing
   \FOR {every model in the \(List_{ML}\)}
   \STATE \redsout{Call $GS$ \& $RS$} $\Rightarrow$ \textcolor{red}{Call Hyp-ABC}
   \STATE { \redsout{\textbf{if} GS.Acc  larger than RS.Acc \textbf{then}}} 
   \STATE \textcolor{red}{{\textbf{if} ABC.ACC larger than reported {(GS \& RS)}.Acc \textbf{then}}}
   \STATE \redsout{Replace model, GS BestHPs}
   \STATE $\Rightarrow$ \textcolor{red}{{Replace Model, ABC BestHPs}}
   \STATE {\textbf{end if}}
   \STATE {\redsout{\textbf{if} RS.Acc  larger than GS.Acc \textbf{then}}}
   \STATE \hspace{10pt}\redsout{Replace model, RS BestHPs}
   \STATE \redsout{\textbf{end if}}
  \ENDFOR
  \RETURN FinalModel, BestHP, Final.Acc, TuningTime
  \end{algorithmic}
  \label{alg:GRS}
  \end{algorithm}
  
Since RF, XGBoost, and SVM are computationally highly expensive regarding tuning time when using automated HPO methods \cite{zahedi2021search}, they are selected to be explored in this study. The tuning time complexity in these three models is mainly due to the model's complexity and the number, the range, and the type of hyper-parameters.

%========================================
\section{Experimental Results}
\label{sec:results}
In this section, we discuss the dataset, evaluation metrics, and experimental results in depth.
\subsection{Dataset}
In this study, we used a real-world educational data, Multiple-Institution Database for Investigating Engineering Longitudinal Development (MIDFIELD)\cite{ohland2004creation}, to predict students' graduation. This dataset is a unit-record longitudinal dataset for undergraduate students from 16 universities. MIDFIELD contains all the information that appears in students' academic records, including demographic data (sex, age, and race/ethnicity) and information about major, enrollment, graduation, and school and pre-school performance. We used a reduced version of MIDFIELD, including only students majoring in computing fields (with CIP=11). This version of the MIDFIELD dataset is used for a binary classification problem and has 4532 samples with 91 features.

\subsection{Metrics for Performance Evaluation}
The evaluation metrics used in this experiment to evaluate the performance of the classification and the proposed framework are accuracy and execution time. Accuracy is the ratio of the number of correct predictions to the total number of predictions. This metric is used since the dataset is balanced and the number of samples belonging to each class is almost equal.

We also used cross-validation to partly reduce or prevent over-fitting. This helped to find a stable optimum that is suitable for all the subsets of the dataset instead of only a singular validation set \cite{hutter2019automated}. Table \ref{tab:cls-abc} summarizes the results and shows the accuracy of different HPO methods among different ML models. It is important to note that cross-validation also increases the tuning time by the number of folds. Therefore, we used 3-fold cross-validation to avoid very high tuning time. However, we parallelized the cross-validation process to reduce the impact of cross-validation on execution time. To extend the experiment, we also separated the MIDFIELD dataset into train and test sets and repeated the experiment. The train set contains 80\% of the dataset, and the test set includes 20\% of the dataset.

In this experiment, the Scikit-learn and XGBoost libraries were used to leverage ML models.
All experiments were conducted using Python 3.8 on Amazon Web Services (AWS) servers with four v-CPUs up to a 3.0 GHz scalable processor and 16 GB RAM.

\begin{figure*}[ht] 
%\centering \caption{Food sources number effect on the accuracy (cv=3)}
%\begin{minipage}{.4\textwidth}
\begin{subfigure}[b]{0.45\textwidth}
\begin{tikzpicture}[scale=0.8]
    \begin{axis}[
        width=0.43\paperwidth,
        height=0.31\paperwidth,
        major x tick style = transparent,
        ymin=87,
        ybar=3*\pgflinewidth,
        bar width=10pt,
        ymajorgrids = true,
        ylabel = {Accuracy (\%)},
        symbolic x coords={SVM,RF,XGBoost},
        xtick=data,
        ytick = {87.5,88,88.5},
        %ytick distance=2,
        %scaled y ticks = true,
        enlarge x limits=0.2,
        legend cell align=center,
        legend style={
                at={(1,0)},
                anchor=south east,
                column sep=1pt,
                nodes={scale=0.7, transform shape},
        }
    ]
        \addplot[style={bblue,fill=bblue,mark=none}]
            coordinates {(SVM,87.93) (RF, 87.57) (XGBoost,87.97)};

        \addplot[style={rred,fill=rred,mark=none}]
             coordinates {(SVM,88.00) (RF, 88.74)  (XGBoost,88.64)};

        \addplot[style={ggreen,fill=ggreen,mark=none}]
             coordinates {(SVM,88.00) (RF, 88.77) (XGBoost,88.84)};
             
        \legend{NP=20,NP=50,NP=100}
    \end{axis}
\end{tikzpicture}
\caption{Effect of number of food sources in HyP-ABC on accuracy}
\label{a}
%\label{fig:pop-cv}
\end{subfigure}
%\end{minipage}
\centering
%\caption{Food sources number effect on the accuracy (80:20 test/train sets)}
%\begin{minipage}{.4\textwidth}
\begin{subfigure}[b]{0.45\textwidth}
\begin{tikzpicture}[scale=0.8]
    \begin{axis}[
        width=0.43\paperwidth,
        height=0.31\paperwidth,
        major x tick style = transparent,
        ymin=0,
        ymax=100,
        ybar=3*\pgflinewidth,
        bar width=10pt,
        ymajorgrids = true,
        ylabel = {Execution Time (Hours)},
        symbolic x coords={GS, RS, HyP-ABC},
        xtick=data,
        ytick = {10,25,40,55,70,85},
        %ytick distance=2,
        %scaled y ticks = true,
        enlarge x limits=0.2,
        legend cell align=center,
        legend style={
                at={(1,0.75)},
                anchor=south east,
                column sep=1pt,
                nodes={scale=0.7, transform shape},
        }
    ]
        \addplot[style={cherryRed,fill=cherryRed,mark=none}]
            coordinates {(GS,97.47) (RS, 26.33) (HyP-ABC,34.15) };

        \addplot[style={emerald,fill=emerald,mark=none}]
            coordinates {(GS,64.55) (RS, 10.85) (HyP-ABC,1.55) };

        \addplot[style={orange,fill=orange,mark=none}]
            coordinates {(GS,68.97) (RS, 29.60) (HyP-ABC,29.22) };
             
        \legend{SVM,RF,XGBoost}
    \end{axis}
\end{tikzpicture}
\caption{Comparison of execution times among HPO methods}
\label{a}
\end{subfigure}
\caption{Comparison of Execution Time and Accuracy Among Different ML Models Using 3-fold Cross Validation}
\label{fig:cv}
%\end{minipage}
\end{figure*}
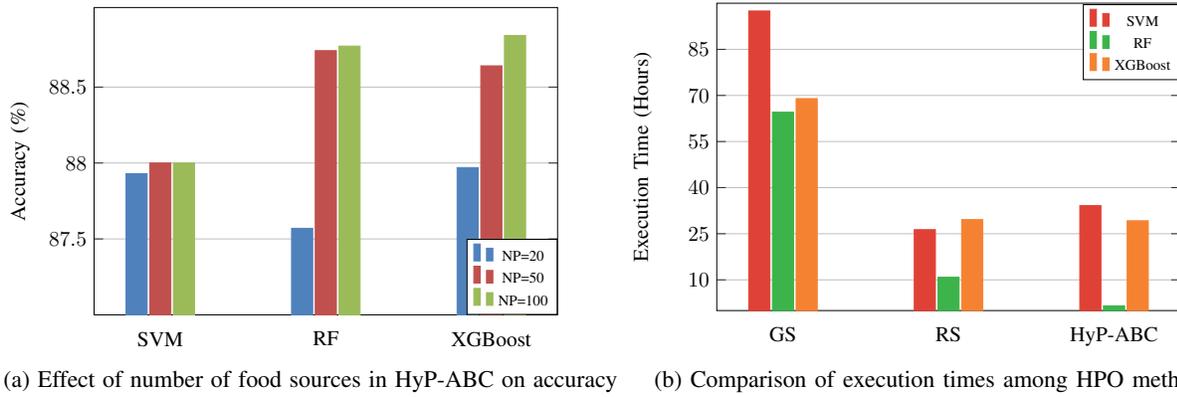

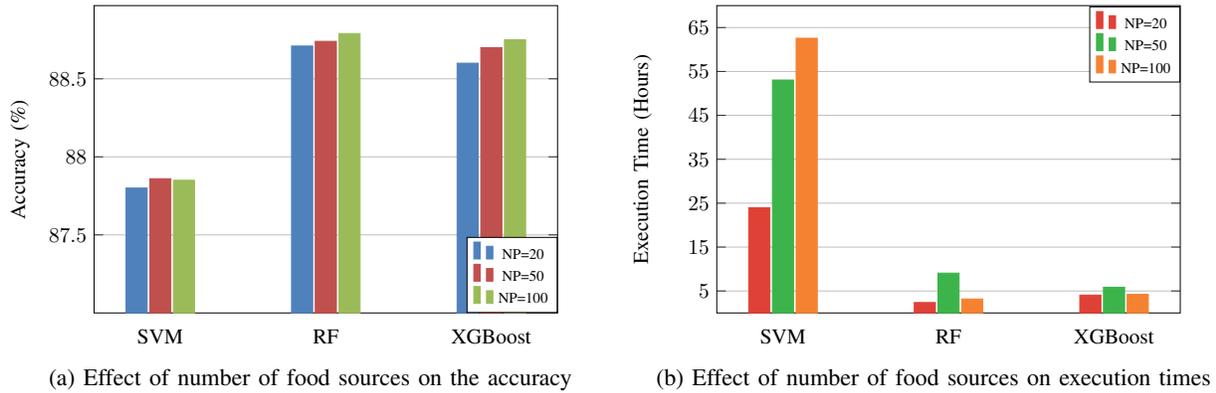
\begin{figure*}[ht] 
%\centering \caption{Food sources number effect on the accuracy (cv=3)}
%\begin{minipage}{.4\textwidth}
\begin{subfigure}[b]{0.45\textwidth}
\begin{tikzpicture}[scale=0.8]
    \begin{axis}[
        width=0.43\paperwidth,
        height=0.31\paperwidth,
        major x tick style = transparent,
        ymin=87,
        ybar=3*\pgflinewidth,
        bar width=10pt,
        ymajorgrids = true,
        ylabel = {Accuracy (\%)},
        symbolic x coords={SVM,RF,XGBoost},
        xtick=data,
        ytick = {87.5,88,88.5},
        %ytick distance=2,
        %scaled y ticks = true,
        enlarge x limits=0.2,
        legend cell align=center,
        legend style={
                at={(1,0)},
                anchor=south east,
                column sep=1pt,
                nodes={scale=0.7, transform shape},
        }
    ]
        \addplot[style={bblue,fill=bblue,mark=none}]
            coordinates {(SVM,87.80) (RF, 88.71) (XGBoost,88.60) };

        \addplot[style={rred,fill=rred,mark=none}]
             coordinates {(SVM,87.86) (RF, 88.74 )  (XGBoost,88.70)};

        \addplot[style={ggreen,fill=ggreen,mark=none}]
             coordinates {(SVM,87.85) (RF,88.79 ) (XGBoost,88.75) };
             
        \legend{NP=20,NP=50,NP=100}
    \end{axis}
\end{tikzpicture}
\caption{Effect of number of food sources on the accuracy}
\label{b}
%\label{fig:pop-ncv}
\end{subfigure}
%\end{minipage}
\centering
%\begin{minipage}{.4\textwidth}
\begin{subfigure}[b]{0.45\textwidth}
\begin{tikzpicture}[scale=0.8]
    \begin{axis}[
        width=0.43\paperwidth,
        height=0.31\paperwidth,
        major x tick style = transparent,
        ymin=0,
        ymax=70,
        ybar=3*\pgflinewidth,
        bar width=10pt,
        ymajorgrids = true,
        ylabel = {Execution Time (Hours)},
        symbolic x coords={SVM,RF, XGBoost},
        xtick=data,
        ytick = {5,15,25,35,45,55,65},
        %ytick distance=2,
        %scaled y ticks = true,
        enlarge x limits=0.2,
        legend cell align=center,
        legend style={
                at={(1,0.75)},
                anchor=south east,
                column sep=1pt,
                nodes={scale=0.7, transform shape},
        }
    ]
        \addplot[style={cherryRed,fill=cherryRed,mark=none}]
            coordinates {(SVM,23.95) (RF, 2.37) (XGBoost,4.07) };

        \addplot[style={emerald,fill=emerald,mark=none}]
            coordinates {(SVM,53.05) (RF, 09.07) (XGBoost,5.85)};

        \addplot[style={orange,fill=orange,mark=none}]
            coordinates {(SVM,62.57) (RF, 3.18) (XGBoost,4.23) };
             
        \legend{NP=20,NP=50,NP=100}
    \end{axis}
\end{tikzpicture}
\caption{Effect of number of food sources on execution times}
\label{b}
%\end{minipage}
\end{subfigure}
\caption{Comparison of Execution Time and Accuracy Among Different ML Models Using a 80:20 Ratio Test/Train Set}
\label{fig:ncv}
\end{figure*}

\begin{table}[b]
\centering
\caption{Performance evaluation of utilizing HPO methods to the ML classifiers on the MIDFIELD dataset (cv=3)}
\label{tab:cls-abc}
%\begin{adjustbox}{width=1.5\columnwidth,center}
\def\arraystretch{1.5}
\begin{tabular}{|c|c|c|c|}
\hline
\textbf{} & \multicolumn{3}{c|}{Accuracy (\%)}\\
\hline
\textbf{HPO Method / ML Classifier} & \textbf{RF}
& \textbf{XGBoost}   & \textbf{\makecell{SVM}}\\
\hline
\textbf{\makecell{GS}}   &  \makecell{$>$ 1 year} & \makecell{$>$ 1 year} & \makecell{87.99}\\
\hline
\textbf{\makecell{GS \cite{zahedi2021search} \\(minimized search space)}}  &  \makecell{88.34} & \makecell{88.33} & \makecell{NA} \\
\hline
\textbf{RS \cite{zahedi2021search}} & \makecell{88.37}  & \makecell{88.80} & \makecell{87.43}\\
\hline
\textbf{HyP-ABC} &  \makecell{\textbf{88.77}} & \makecell{\textbf{88.84}} & \makecell{\textbf{88.00}}\\
\hline
\end{tabular}
%\end{adjustbox}
\end{table} 

\subsection{Results}
The above-described HyP-ABC is developed to achieve the best hyper-parameters of SVM, RF, and XGBoost algorithms in an efficient time. Figure \ref{fig:cv} shows the comparison of tuning time and accuracy among the three ML models. Figure \ref{fig:cv}(a) shows the impact of the different numbers of population on the accuracy; As can be seen, the more the number of food sources, the better the accuracy. Figure \ref{fig:cv}(b) also shows that the execution time for all the ML models decreased in comparison to the GS method. Regarding RS, the execution time has also decreased or did not change except for the SVM model. The slight differences in tuning time for RS and HyP-ABC are mainly due to the maximum number of evaluations equal that is defined as equal to the number of iterations in RS. That being said, even if execution time remained stagnant for some cases but the accuracy of the model has been improved, which is promising. It is important to note that for RF and XGBoost algorithms, the GS tuning time considering all the configurations leads to a tuning time of more than years. This estimation was calculated from the average tuning time for each configuration from the RS experiment. Therefore the experiment for the GS method is implemented using a reduced search space with steps of five to make the experiment feasible.

As for the performance of the proposed method, the overall accuracy of all the ML models has increased after hyper-parameter tuning using HyP-ABC. Therefore, the HyP-ABC method outperformed the previous HPO methods from the most recent work, and more importantly, the tuning time has decreased. This is due to the fact that the ABC method in general, unlike GS and RS, is a model-based method and has a methodical way of moving toward the close to an optimal solution. Also, among all the models, XGBoost had the best accuracy score. Hence, XGBoost is the final model to be used on the MIDFIELD dataset in this study.

\section{Conclusion and Discussions}
\label{sec:conclusion}
%========================================
In this paper, a modified evolutionary optimization algorithms for hyper-parameter tuning in ML models, referred to as HyP-ABC, was proposed to enable hyper-parameter tuning of the conventional ML classifiers. We carried out the experiment using the MIDFIELD dataset. The behavior of HyP-ABC has been explored under a different number of populations, and the accuracy and execution time of the HyP-ABC has been compared with HPO methods in an earlier study.

Experimental results verified that HyP-ABC outperforms the tuning methods used in the state-of-the-art methods,  %recent studies \cite{zahedi2021search,zahedi2020leveraging}; 
 Hyp-ABC improves the classification accuracy, and more importantly, it decreases the tuning time significantly. HyP-ABC can be deployed to solve the HPO problems with large search spaces. The major advantage of this work is the strong potential to improve tuning time without negatively affecting the performance. 
To summarize, HyP-ABC is recommended for optimizing RF,  GXBoost, and SVM. In the future work, we will explore the semantic initialization of the population to further reduce the tuning time. Also, time complexity of algorithms such as SVM is highly dependent on the number of features and samples of a dataset. % However further investigation is required to find the optimal or near-optimal hyper-parameters even with a very limited budget. Hence their function evaluation can be very slow for large-scale datasets. Exploration of feature selection and data reduction techniques is encouraged for future work.

%========================================
%========================================
\section{Acknowledgement}
\label{sec:acknowledge}
%========================================
The authors would like to thank Dr. Matthew Ohland from Purdue University for giving access to the MIDFIELD dataset, and Dr. Monique Ross from Florida International University for her valuable inputs.

%\section*{Acknowledgment}



%\bibliographystyle{IEEEtran}

% \newpage

% \end{thebibliography}
\bibliographystyle{IEEEtran}
%\clearpage

\bibliography{IEEEabrv,myBib.bib}

\end{document}